\documentclass[conference]{IEEEtran}
\IEEEoverridecommandlockouts
\usepackage{cite}
\usepackage{amsmath,amssymb,amsfonts}
\usepackage{algorithmic}
\usepackage{graphicx}
\usepackage{textcomp}
\usepackage{xcolor}
\usepackage{subcaption}
\usepackage{booktabs}
\usepackage{siunitx}
\usepackage{multirow}
\usepackage{amsmath,amssymb,amsfonts}
\usepackage[ruled,vlined]{algorithm2e}
\SetAlFnt{\small}

\renewcommand\baselinestretch{0.96}

\newenvironment{Operation Flow}[1][htb]
  {
   \begin{algorithm}[#1]%
  }{\end{algorithm}}

\newcommand{\floor}[1]{\left\lfloor #1 \right\rfloor}

\def\BibTeX{{\rm B\kern-.05em{\sc i\kern-.025em b}\kern-.08em
    T\kern-.1667em\lower.7ex\hbox{E}\kern-.125emX}}
\begin{document}

\makeatletter
\def\ps@IEEEtitlepagestyle{
  \def\@oddfoot{\mycopyrightnotice}
  \def\@evenfoot{}
}
\def\mycopyrightnotice{
  {\footnotesize
  \begin{minipage}{\textwidth}
  \centering
 \copyright~2021 IEEE.  Personal use of this material is permitted. Permission from IEEE must be 
obtained for all other uses, in any current or future media, including 
reprinting/republishing this material for advertising or promotional purposes, creating new 
collective works, for resale or redistribution to servers or lists, or reuse of any copyrighted 
component of this work in other works.
  \end{minipage}
  }
}

\title{In-Hardware Learning of Multilayer Spiking Neural Networks on a Neuromorphic Processor}


\author{\IEEEauthorblockN{Amar Shrestha, Haowen Fang, Daniel Patrick Rider, Zaidao Mei and Qinru Qiu}
\IEEEauthorblockA{\textit{Department of EECS, Syracuse University, Syracuse, USA}\\
\text{\{amshrest, hfang02, dprider, zmei05, qiqiu\}@syr.edu}}
}



\maketitle

\begin{abstract}
Although widely used in machine learning, backpropagation cannot directly be applied to SNN training and is not feasible on a neuromorphic processor that emulates biological neuron and synapses. This work presents a spike-based backpropagation algorithm with biological plausible local update rules and adapts it to fit the constraint in a neuromorphic hardware. The algorithm is implemented on Intel’s Loihi chip enabling low power in-hardware supervised online learning of multilayered SNNs for mobile applications. We test this implementation on MNIST, Fashion-MNIST, CIFAR-10 and MSTAR datasets with promising performance and energy-efficiency, and demonstrate a possibility of incremental online learning with the implementation.
\end{abstract}


\section{Introduction}

Emerging neuromorphic hardware that adopts the event driven behavior of spiking neural networks (SNN) has outstanding energy efficiency and is believed to be effective for edge computing or working with certain type of sensor, such as dynamic vision sensor (DVS), whose output is sparse by nature \cite{son20174, liu2019event, haessig2019neuromorphic}. However, the lack of a unified robust learning algorithm limits the SNN to shallow networks with low accuracies. The commonly accepted synaptic plasticity rule, the Hebbian rule, is an unsupervised learning, which cannot be directly applied to solve vast majority of machine learning problems that require supervised learning. A common approach is to train an Artificial neural network (ANN) and convert it into SNN \cite{diehl2015fast, esser2015backpropagation}, however, this requires the training to be performed offline.

In-hardware learning is the first step towards building an autonomous machine that learns continuously from its environment and experiences. It means performing inference and learning using the same model running on the same hardware platform. It provides the ability to compensate any device variation and/or environment noise in the inference stage and a possibility of incremental learning when new classes become available to the system being deployed. However, due to limited storage capacity of low cost edge devices, the training data for
real life in-hardware learning is streamed instead of batched. Thus, the learning must be online. This makes in-hardware learning a more significant challenge than offline learning.


Backpropagation, a gradient-based optimization algorithm, is a standard training technique for ANNs. However, it cannot be directly applied to the in-hardware learning of an SNN running on a neuromorphic processor  due to several reasons; (1) spiking neuron’s activities are not differentiable, 
(2) the connections between neurons in SNN are unidirectional such that a backward path must be added explicitly with constantly updated weights during learning 
, (3) errors in ANN are propagated as real values and (4) weight update of a synapse is not dependent only on locally available information as required in a neuromorphic hardware \cite{Davies2018}.

Most works approximating backpropagation for SNNs have some  limitations. Some require neurons to have high-precision backpropagating error derivatives \cite{wu2018spatio, tavanaei2019bp, o2016deep}. In some other works, neuron must know the membrane potential of its presynaptic neighbors in order to determine the synaptic weight change \cite{jin2018hybrid}. In \cite{tavanaei2019bp, o2016deep, neftci2017event}, each neuron must also know the error derivatives of its postsynaptic neighbors in order to calculate its own error derivative. All these limitations violate the constraints of local communication rule in spike domain and require additional hardware to support the backpropagation of real valued error derivatives and communication of membrane potentials. Meanwhile, approaches \cite{Severa2018}\cite{Rasmussen2018}, which convert trained ANNs to SNNs, are not conducive to neuromorphic implementation.

To the best of our knowledge, the recent Error-Modulated Spike-timing-dependent plasticity (EMSTDP) algorithm
\cite{Shrestha2019} is the only supervised learning model that applies the same type of integrate and fire (IF) neuron in the forward and backpropagation path. The algorithm enhances the biological plausibility of backpropagation by introducing a weight update rule that resembles the rate-based STDP \cite{Bengio2015}. However, there is still a gap between the EMSTDP algorithm and the in-hardware online learning on a neuromorphic hardware. First of all, EMSTDP assumes arbitrary data precision in weight coefficients and unlimited neuron and synapse resources. Secondly, the weight update rule of EMSTDP and the error modulation process still require special hardware support. Finally, many of the existing data sets were designed for ANN. The inputs are vectors of real numbers. Traditional rate coding that represents the real number as a sequence of spike trains will introduce significant amount of I/O activities. In this work, we bridge the gap and adapt EMSTDP to fit the constraints of a neuromorphic hardware and implement it on Intel’s Loihi chip. The following four approximation techniques were introduced that enable the neuromorphic hardware implementation of the EMSTDP :
\begin{itemize}
  \item Multi-compartment neuron were adopted in the feedback path where the error spikes can be gated by the activities of the feedforward neurons.
  \item The change in the post-synaptic spiking rate is approximated by using the built in post-synaptic trace counter and a two-phase operation.
  \item Direct feedback alignment is adopted to significantly reduce the number of neurons in the feedback path and hence lower hardware cost and improve learning accuracy.
  \item Reduce the I/O activities by programming the bias of the input neurons using the real valued inputs and generating the spike sequences inside the chip. 
\end{itemize}

The cost and accuracy impact of these techniques will be discussed in the paper. This is the first work of fully spike-based online in-hardware supervised learning on a neuromorphic hardware. The implementation is tested on MNIST, Fashion-MNIST, CIFAR-10 and MSTAR datasets with promising performance and energy-efficiency. We also demonstrate a possibility of incremental online learning with the implementation.


\section{Background}



\subsection{Error-Modulated STDP}\label{lab:EMSTDP}

\cite{Shrestha2019} extends the backpropagation algorithm for SNNs through a suitable sequence of approximation techniques. First, the forward and the backward paths are separated into separate networks with spiking neurons. Simple Integrate-and-Fire (IF) neurons are used in both paths. For the forward path, the membrane potential of neurons in layer $i$ at time $t$ is:

\vspace{-3mm}
\begin{equation} \label{eq:mempot_original}
U_i (t) =  \sum w_{i-1,i}.s_{i-1}(t) + b_i + U_i (t - 1)
\end{equation}
\vspace{-3mm}
$$\textrm{if} \quad U_i(t) \geq \theta \quad \textrm{then} \quad s_i(t) = 1, \quad \textrm{else} \quad s_i(t) = 0$$

where $w_{i-1,i}$ is the synaptic weight  between pre-synaptic neurons in layer $i-1$  and post-synaptic neurons in layer $i$, $b_i$ is the bias. $s_{i-1}(t)$ and $s_i(t)$ are pre and postsynaptic spike trains respectively. The neuron spikes when the membrane potential reaches the threshold $\theta$ and is reset to the resting potential of $0$. Its activation function is approximated in terms of the spike count $h,$ accumulated sub-threshold membrane potential $u$, and spiking threshold $\theta$ over a duration $T$:
\vspace{-1mm}
\begin{equation}\label{eq:spikecount}
h=f(u)= \floor{\frac{u}{\theta}}
\end{equation}

It's derivative is approximated as that of a shifted ReLU.


$\epsilon_i$ represents the backpropagated error derivatives of the $i_{th}$ layer, based on the backpropagation algorithm such that
\vspace{-2mm}
\begin{equation}\label{eq:epsilon}
     \epsilon_i = \epsilon_{i+1} \cdot h'_{i+1} \cdot w_{i,i+1}^T 
\end{equation}
\vspace{-5mm}

And error spike count $e$ in the error path is represented as:
\vspace{-2mm}
\begin{equation}\label{eq:e_equal_epsilon}
     e_{i+1} = \epsilon_{i+1} \cdot h'_{i+1} 
\end{equation}
\vspace{-5mm}


Such that,
\vspace{-2mm}
\begin{equation}\label{eq:epsilon_to_e}
     \epsilon_i = e_{i+1} \cdot w_{i,i+1}^T
\end{equation}

This means that we can represent the error derivative also using spikes, and use the same IF neurons (without bias) given by (\ref{eq:mempot_original}) in the feedback path. $\epsilon_i$ is the accumulated sub-threshold membrane potential of the error neuron whose input is the spikes  $e_{i+1}$ from the upper-level error neuron. 
In addition to the threshold based firing, the output of the neurons in the feedback path is also gated by the $h'_i$, which is a constant when the neuron in the corresponding feedforward layer has output activities and zero otherwise.  


L2 loss's derivative is entirely spike-based. To represent that, the accumulated sub-threshold membrane potential of neurons in the first layer in the feedback path is calculated as
\begin{equation}\label{eq:loss}
     \epsilon_l = w_L \hat{h_l} + (-w_L)h_l
\end{equation}
where $\hat{h}$ and $h$ are target and predicted spike trains.

Local learning is achieved through the idea of target propagation. In \cite{Lee2015} , the targets are determined utilizing an autoencoder with inverse synaptic weights which are learnt through reconstruction. In EMSTDP, the error derivatives are used as the difference to determine the target spike count $\hat{h}_{l}$. Such that the local weight update is formulated as
\vspace{-1mm}
\begin{equation}\label{eq:weight_update}
\Delta w_{i-1, i}=\eta \cdot\left(\widehat{h_{l}}-h_{i}\right) \cdot h_{i-1}
\end{equation}
where $\eta$ is the learning rate. This computation is done in two phases: first phase to compute $h_{i},$ and second phase to compute $\hat{h}_{\imath}$ and perform the weight update.
Since, the weight is constantly changing during learning, maintaining the coherence between the two copies will create significant overhead. EMSTDP adopts the Feedback Alignment (FA) method \cite{Lillicrap2016}, to relax this constraint. Instead of maintaining an exact copy of the weight in both the feedback and feedforward path, a random weight matrix is used in the feedback path.



\begin{figure*}[t]
    \centering
    \begin{subfigure}{0.99\textwidth}
        \centering
        \begin{subfigure}{0.33\textwidth}
        \centering
        \includegraphics[height=4.5cm]{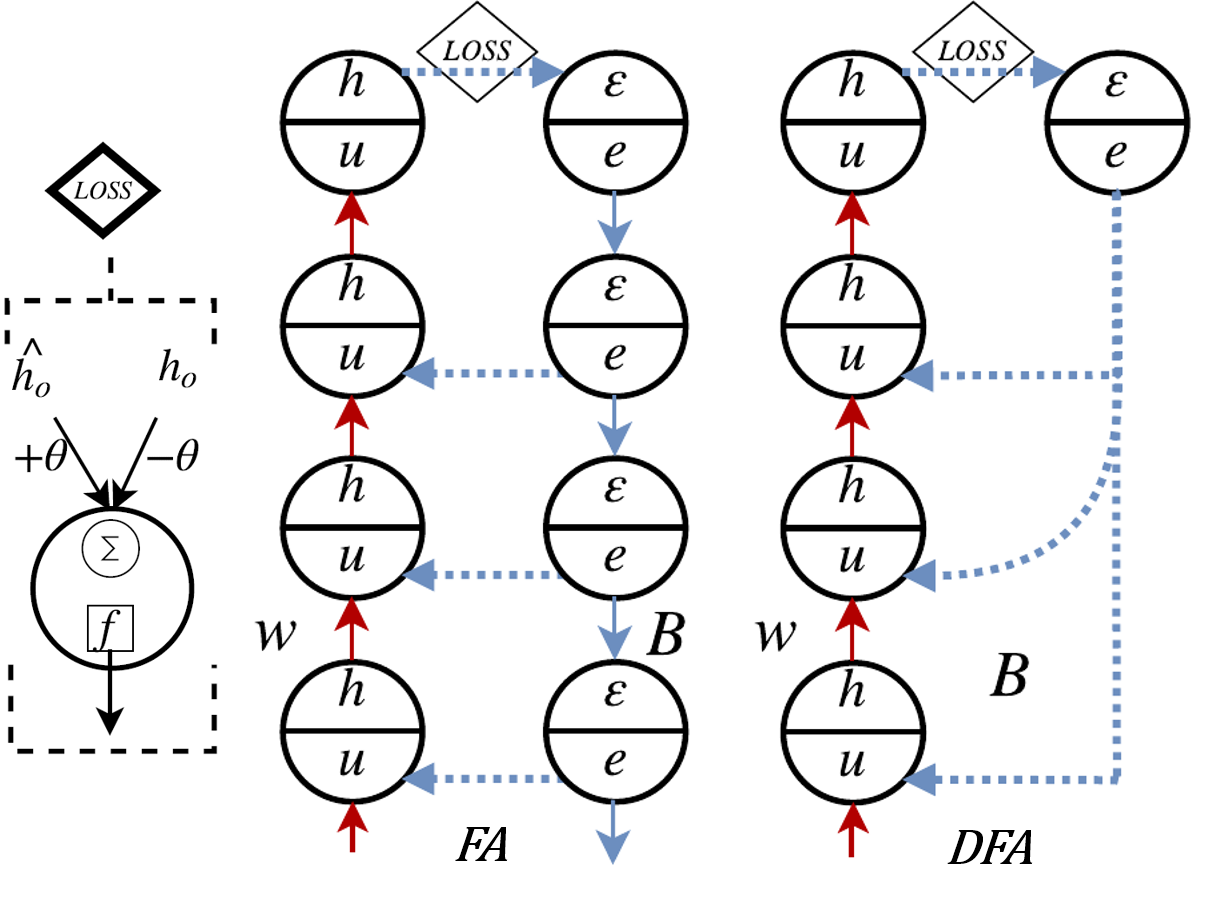}
        \caption{}
        \label{fig:emstdp}
    \end{subfigure}
        \centering
    \begin{subfigure}{0.23\textwidth}
        \includegraphics[height=4.0cm]{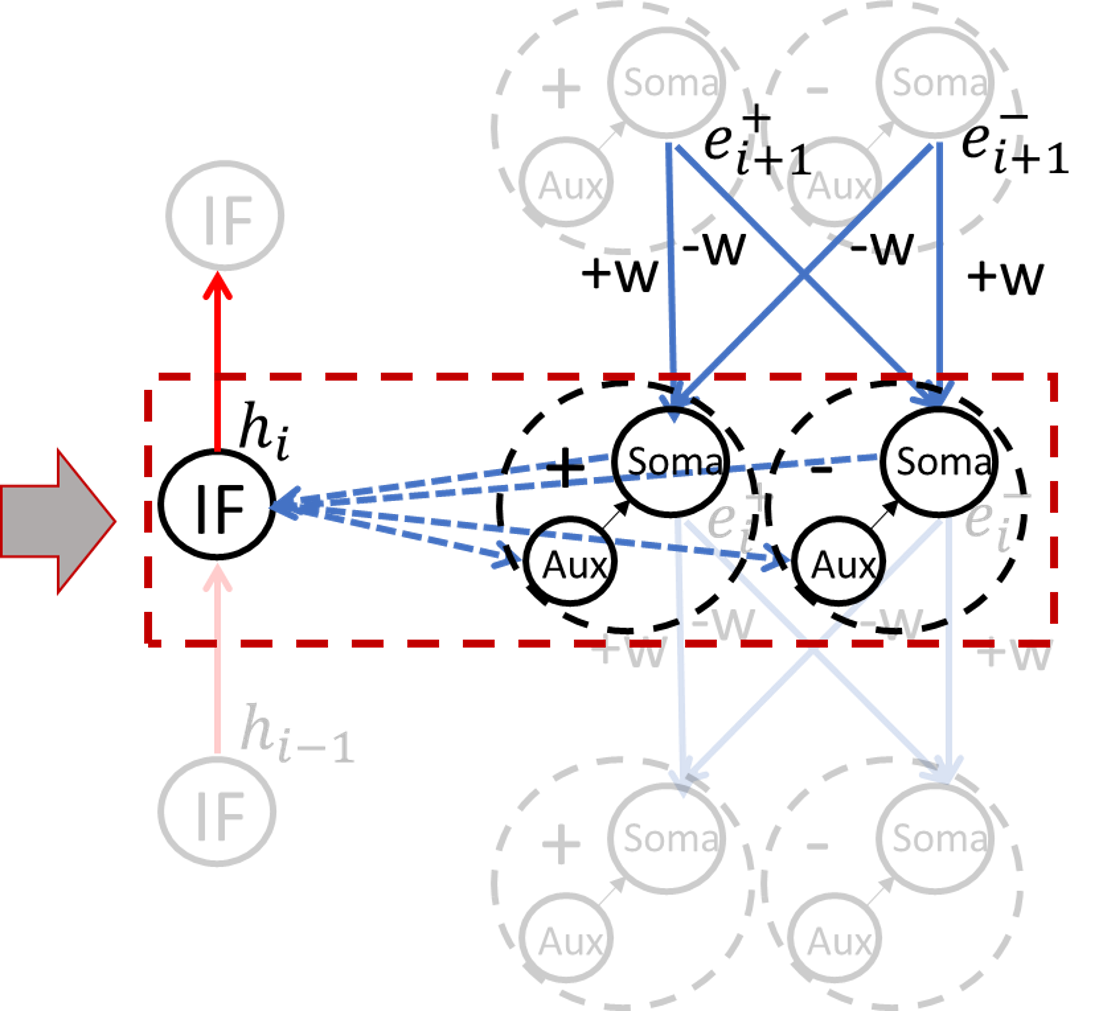}
        \caption{}
        \label{fig:loihi_network}
    \end{subfigure}
        \centering
    \begin{subfigure}{0.41\textwidth}
        \centering
        \includegraphics[height=4.0cm]{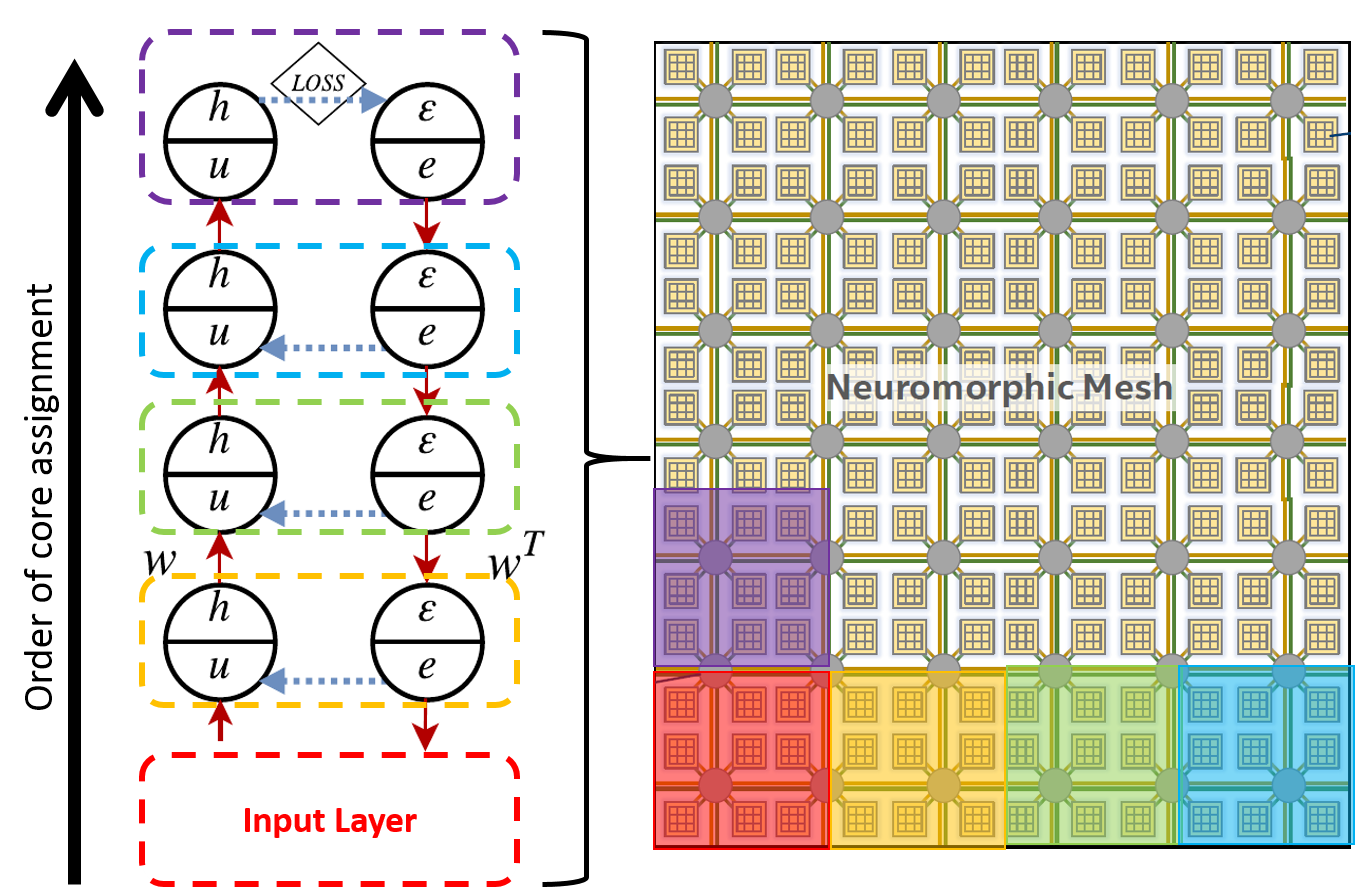}
        \caption{}
        \label{fig:loihi_mapping}
    \end{subfigure}
\end{subfigure}
\vspace{-0.3cm}
\caption{(a) EMSTDP - Network (b) Loihi - Network (c) Loihi - Mapping }
\label{fig:Loihi_network_mapping}
\vspace{-5mm}
\end{figure*}

  

\subsection{Synaptic Plasticity in Loihi}\label{lab:Loihi}

Loihi \cite{Davies2018} is a digital neuromorphic chip recently developed by Intel. Loihi's 128-neuromorphic cores implement 130,000 artificial CUBA leaky-
integrate-and-fire neurons and 130 million synapses. 
The CUBA neurons have
two internal state variables; synaptic response current $u_{i}(t)$ and membrane potential $v_{i}(t)$. The synaptic response current is the decaying weighted incoming spikes and the membrane potential is formulated as:
\vspace{-3mm}
\begin{equation}\label{eq:loihi_mem_pot}
\dot{v}_{\imath}(t)=-\frac{1}{\tau_{v}} v_{i}(t)+u_{i}(t)
\end{equation}

Here, the integration is leaky as captured by the time constant $\tau_{v}$ . The neuron sends out a spike when its membrane potential passes its firing threshold $\theta_{i}$ and is reset to 0 right after. 
The neuron parameters are tunable thus they can be adapted to an IF neuron and other more complicated behaviors.

Loihi also provides a programmable microcode learning engine for on-chip SNN training. 
In Loihi, each synapse is associated with integer-valued synaptic variables and multiple presynaptic traces, and whereas
compartment with postsynaptic traces. 
SNN synaptic weight adaptation rules must satisfy a locality constraint: each weight can only be
accessed and modified by the destination neuron, and the update can only depend on locally available
information, such as the spike trains from the presynaptic (source) and postsynaptic (destination)
neurons. 
The functional form of adaptation rules to apply is described in sum-of-products form:
\vspace{-3mm}
\begin{equation}\label{eq:loihi_sop_form}
z:=z+\sum_{i=1}^{N_{p}} S_{i} \prod_{j=1}^{n_{i}}\left(V_{i, j}+C_{i, j}\right)
\end{equation}
\vspace{-3mm}

where $z$ is the transformed synaptic variable (weight, delay or tag), $V_{i, j}$ refers to some choice of
input variable (synaptic variables and traces) available to the learning engine, and $C_{i, j}$ and $S_{i}$ are microcode-specified signed constants. Regular pairwise and triplet STDP rules can be implemented along with more complicated adaptation rules utilizing this form.

Loihi has been shown to produce impressive performance and energy efficiency for various tasks both for inference-only and in-hardware learning tasks. However, there is still a lack of an online supervised learning algorithm to train multilayered SNNs in Loihi due to the implicit constraints relating to the neuromorphic hardware such as precision, connectivity across cores etc and also the constrained functional form for the learning rule.

\section{Adapting EMSTDP for in-hardware online learning}\label{sec:implementation}

\subsection{Forward and Error Path}\label{lab:ForErrPath}

Since the synaptic connection between neurons in Loihi is directional, we have to keep two copies of networks, one for forward pass and the other to backpropagate errors. And as mentioned in Section \ref{lab:EMSTDP} and as keeping updated weights in the error path is not viable, we utilize FA weights $B$. As shown in Fig. \ref{fig:loihi_network}, both networks are built on IF neurons.
We simply configure the LIF neuron in Loihi into an IF neuron for the forward path. For that we utilize the maximum time constant $\tau_{v}$ in (\ref{eq:loihi_mem_pot}) such that the membrane potential doesn't leak over time whereas the current decays immediately.

In order to represent both positive and negative errors in the error path, we utilize two channels of spiking neurons. Thus, its necessary to cross connect between positive and negative channels of the source and target layers as shown in Fig. \ref{fig:loihi_network}. In both channels, the spike rate represents the absolute value of the error. 
Let $\epsilon_{i}^{+}$ and $\epsilon_{i}^{-}$ represents the accumulated sub-threshold membrane potential in the positive and negative channels in the $i_{th}$ layer. Following (\ref{eq:epsilon_to_e}), they can be calculated as the following: 

\vspace{-3mm}
\begin{equation}\label{eq:weight_cross}
\begin{cases}
     \epsilon_{i}^{+} = e_{i+1}^{+} \cdot w_{i,i+1}^T + e_{i+1}^{-} \cdot (-w_{i,i+1}^T) \\
     \epsilon_{i}^{-} = e_{i+1}^{+} \cdot (-w_{i,i+1}^T) + e_{i+1}^{-} \cdot w_{i,i+1}^T
\end{cases}
\end{equation}



Additionally, we have to consider the $h’$, i.e. the derivative of neuron activation function in the forward path to gate the output activities in the feedback path to as descried by (\ref{eq:e_equal_epsilon}) and (\ref{eq:epsilon_to_e}). Since $h’$ is either 0 or a constant value, this can be implemented utilizing an AND function in the case of bit streams. We utilize the multi-compartment neurons in Loihi for that purpose. 
Two-compartment neurons with a soma compartment and a corresponding auxiliary compartment are set up for the error path such that the spiking activity of the soma is an AND function of the activity of the soma and the auxiliary compartment. 
The setup of the forward and error path is shown in Fig.\ref{fig:loihi_network}. 




The original EMSTDP with FA has a one-to-one correspondence in the feedback and feedforward neurons. It does not only double hardware requirement but also degrade the learning quality. As the error propagated through layers, the quantization errors accumulated. In this work we further adopt the direct feedback alignment (DFA) \cite{nokland2016direct} from EMSTDP. With DFA, we broadcast the error spike directly from the spike-based loss function in \eqref{eq:loss} with uniformly distributed random weights whereas FA requires an error network through which the error spikes propagate down. Fig. \ref{fig:emstdp} compares the networks using FA and DFA. Compared to FA, the DFA does not only eliminate the neurons on the feedback path, the number of connections on the feedback path is also reduced. This is because the FA connects the neurons in the corresponding hidden layers in the forward and feedback paths, while the DFA directly connects the error neurons in output layer to the hidden layers in the forward path. Because the dimension of the output layer is usually smaller than the hidden layer, the weight matrix of DFA is much smaller than that of the FA. Therefore, DFA not only reduces the number of compartments and neuron cores used in the chip, but also reduces the number of synapses and thus the amount of memory utilized by the synapses in the cores.

\begin{figure}[t]
  \begin{subfigure}{0.99\linewidth}
      \centering
        \begin{subfigure}{0.49\textwidth}
        \centering
      \includegraphics[height=2.0cm,keepaspectratio]{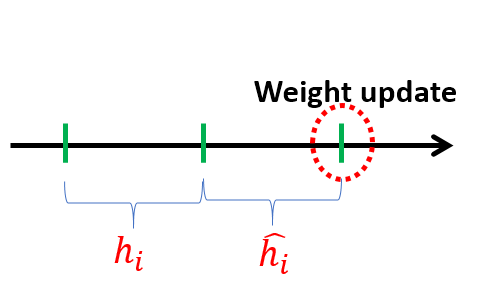}
      \vspace{-2mm}
      \caption{}
      \label{fig:learning1}
      \end{subfigure}
      \hfill
      \centering
        \begin{subfigure}{0.49\textwidth}
        \centering
      \includegraphics[height=2.0cm,keepaspectratio]{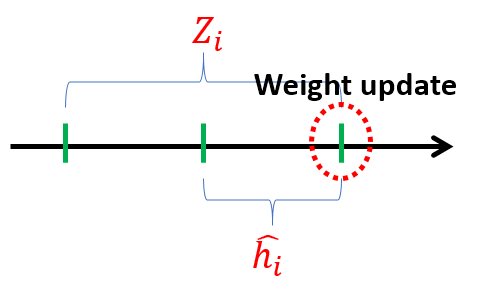}
      \vspace{-2mm}
      \caption{}
      \label{fig:learning2}
      \end{subfigure}
  \end{subfigure}
  \caption{EMSTDP weight update (a) Original (b) Loihi}
  \vspace{-5mm}
\end{figure}


\subsection{Learning}

EMSTDP is set up to enable local learning. 
We need to calculate the product of the change in postsynaptic activity and the presynaptic activity as given by (\ref{eq:weight_update}). The change in postsynaptic activity is computed through the difference of target spike count $\hat{h}_{i}$ and the original spike count $h_{i}$, thus, requiring a two-phase operation occurring in a window of time $2T$. In Phase 1 (time $0$ to $T$), the neurons on the forward path respond to the input, and settle down at a specific spiking rate $h$. In Phase 2 (time $T$ to $2T$), the neurons in the backward path pass the errors to the neurons in the corresponding layers in the forward path. The errors “correct” the behavior of those feed forward neurons, and drive their spiking rate to $\hat{h}$. All the operations in the network with an arbitrary number of layer is asynchronous except for the weight update which is computed only at the end of $2T$.

As weight update rule should be tailored to a sum of products form utilizing variables satisfying the locality constraint, EMSTDP update rule in (\ref{eq:weight_update}) can simply be represented as 
\vspace{-2mm}
\begin{equation}\label{eq:weight_update_sop}
\Delta w_{i-1, i}=\eta \cdot \widehat{h_{l}} \cdot h_{i-1}-\eta \cdot h_{i} \cdot h_{i-1}
\end{equation}
\vspace{-5mm}

However, this representation requires preserving information from the first phase $h_{i}$ to use at the end of the second phase as shown in Fig. \ref{fig:learning1}. Loihi has no means to save this particular information for later use. Thus, we modify (\ref{eq:weight_update_sop}) such that all the information is available at the end of second phase when update is performed as shown in Fig. \ref{fig:learning2}.
\vspace{-2mm}
\begin{equation}\label{eq:loihi_weight_update}
\Delta w_{i-1, i}=2 . \eta \cdot \widehat{h_{l}} \cdot h_{i-1}-\eta \cdot Z_{i} \cdot h_{i-1}
\end{equation}
\vspace{-5mm}

Where $\mathrm{Z}_{i}=\widehat{h}_{i}+h_{i} .$ In Loihi, we represent presynaptic spike count $h_{i-1}$ by the pre-trace, postsynaptic target spike count $\widehat{h}_{i}$ by the post-trace and $\mathrm{Z}_{i}$ by the available tag synaptic variable.

\subsection{Mapping to cores}\label{sec:map}

The core-based architecture of Loihi introduces the limitations of fan-ins and fan-outs which is not always reasonable for larger networks. Thus, it’s necessary to develop mapping algorithm to deal with those constraints. Here, we utilize a simple mapping algorithm where the neurons are mapped incrementally onto the cores satisfying the constraints a layer at a time as shown in Fig.1(c) and Operation Flow \ref{algo:running}. For this, we first generate the adjacency matrices for the connectivity between adjacent layers (convolution and dense). This provides the number of fan-ins and fan-outs for each neuron which is used to assign the number of neurons per core.
This method not only help ensure that the constraints are in check but also utilizes the cores optimally by managing the trade-off (Section \ref{sec:power}) between the execution time and the power usage.

\subsection{Running EMSTDP on Loihi}

Operation Flow \ref{algo:running} shows the entire process of running the EMSTDP algorithm on Intel's Loihi. After creating the network $N$ in Intel Loihi's SDK, we initialize the weights $W$ and the random fixed weights $B$ sampled from a uniform distribution and then quantize and scale them to 8 bit integers to fit Loihi's synapses. Now, we have a network which can be mapped into Loihi which is done during compilation and then deployed.

During runtime, in each epoch during training, for each sample $x$, the values are quantized to the length of the phase $T$. This allows for rate coding the input over the phase length. Normally, inputs are provided through rate-coded spikes directly. Each input data corresponds to a sequence spikes at corresponding rate. Each spike insertion requires a communication between the host and the chip, thus a significant overhead. Instead of inserting spikes directly we program the the biases of the input layer neuron, such that $U_{in}\left(t\right)=\ U_{in}\left(t-1\right)+i,\ 0\le t\le T$, where $i$ is the input and $T$ is the phase length. The accumulated membrane potential can be calculated as $u_{in}=\sum_{t=1}^{T}{U_{in}\left(t\right)}=i\ast T$ and the spiking rate of the input neuron is $h_{in}=\left\lfloor\frac{u_{in}}{\theta}\right\rfloor$ which is linearly proportional to input $i$. In other words, this allows the input layer to integrate the bias over the time thus producing spikes with the rate directly proportional to the bias. Using this setup, we need to communicate with the chip only once for every input sample. We insert the label as bias as well to the label neurons. The network then goes through the two phases of the EMSTDP algorithm, updates the weights and then resets networks states (membrane potential and traces) to 0. 




\begin{Operation Flow}[t]
 \caption{In-Hardware Learning on Loihi}
 \label{algo:running}
\SetAlgoLined
 \textbf{Epoch} $E$, \textbf{Training set} $X$, \textbf{Testing set} $Y$, \textbf{Network} $N$\; 
 \textbf{Trainable Layers} $L$ with $l_n$ neurons, \textbf{Phase Length} $T$\;
 \textbf{Create Network} $N$\;
 \textbf{Initialize $N$ and Quantize to 8 bits}: $W$, $B$\;
 Compile and Deploy Network $N$ on Loihi\;
 \For{\textbf{each} $l \in L$}{
   Build $l-1:l$ adjacency matrix\;
   Compute $l_m$\: Optimal number of neurons per core for $l$\; 
   Map $l_n$ to $\frac{l_n}{l_m}$ cores\;
   }
 \For{$epoch \gets 0$ \KwTo $E$}{
  \For{\textbf{each} $x \in X$}{
   Quantize $x$ to $T$ bins\;
   Set input and label bias: $x$\;
   EMSTDP Phase 1: $T$ time steps\;
   EMSTDP Phase 2: $T$ time steps\;
   Update $W$\;
   Reset network state\;
   }
 }
   \For{\textbf{each} $y \in Y$}{
   Quantize $y$ to $T$ bins\;
   Set input bias: $y$\;
   EMSTDP Phase 1: $T$ time steps\;
   Evaluate\;
   Reset network state\;
   }
   
\end{Operation Flow}

\section{Experiments}

\subsection{Online Learning} \label{sec: OL}

In this experiment we evaluate online learning using EMSTDP algorithm on Loihi. In real life applications of in-hardware learning, the training data is received as a stream, and training must be carried out in real-time as the data being received. Techniques such as batch learning, data augmentation are not feasible in such online learning scenario. 

For all the experiments we use a network with structure $W \times H \times C-5 \times 5k \times 16c2s-3 \times 3k \times 8c2s-100d-10d$ ($W$: width, $H$: height, $C$: channel of the input, $k$: kernel size, $c$: number of filters, $s$: stride size, $d$: dense layer), phase length $T$ of 64 time steps and learning rate $\eta = 2^{-3}$.
Here, the convolutional layers are pretrained offline with their respective datasets before mapping on to Loihi whereas the dense layers are trained from scratch in the Loihi. We utilize only two convolutional layers to ease mapping and minimize its influence on the measurements during training of the dense layers. This introduces opportunities of transfer learning when training such convolutional layers in-hardware is not viable. 

In the experiments, we compare the learning performance of EMSTDP running on Loihi with its software (Python) implementation with batch size 1 and full precision weights. 



\begin{table}[t]
\centering
\caption{Performance}
\label{tab:Performance}
\vspace{-2mm}
\begin{tabular}{|c|c|c|c|c|c|}
\hline
\multirow{2}{*}{} & \textbf{Loihi} &
 \begin{tabular}[c]{@{}c@{}}\textbf{Python}\\ \textbf{(FP)}\end{tabular} & 
 \textbf{Loihi} & 
 \begin{tabular}[c]{@{}c@{}}\textbf{Python}\\ \textbf{(FP)}\end{tabular}  \\ \hline \hline
 \textbf{MNIST}
 & 94.5\% 
 & 98.9\% 
 & 94.7\% 
 & 98.9\% \\ 
 \textbf{Fashion-MNIST}
 & 84.3\% 
 & 92.7\% 
 & 84.8\% 
 & 92.5\% \\ 
 \textbf{MSTAR (10 class)}
 & 78.4\% 
 & 83.5\% 
 & 79.5\% 
 & 83.3\% \\ 
 \textbf{CIFAR10}
 & 61.6\% 
 & 64.2\% 
 & 62.2\% 
 & 64.4\% \\ 
\hline \hline
\multicolumn{1}{|l|}{} & 
\multicolumn{2}{c|}{\textbf{FA}} & \multicolumn{2}{c|}{\textbf{DFA}} \\ \hline
\end{tabular}
\vspace{-3mm}
\end{table}

\begin{table}[t]
\centering
\caption{Power and Energy}
\label{tab:Power}
\vspace{-0.2cm}
\begin{tabular}{|c|c|c|c|c|c|c|} 
\hline 
\multirow{2}{*}{} & \textbf{FPS} &
\begin{tabular}[c]{@{}c@{}}\textbf{Power}\\ \textbf{(W)}\end{tabular} &
 \begin{tabular}[c]{@{}c@{}}\textbf{Energy}\\ \textbf{(mJ/img)}\end{tabular} & 
 \textbf{FPS} & 
\begin{tabular}[c]{@{}c@{}}\textbf{Power}\\ \textbf{(W)}\end{tabular} &
 \begin{tabular}[c]{@{}c@{}}\textbf{Energy}\\ \textbf{(mJ/img)}\end{tabular}  \\
\hline \hline
 \textbf{i7 8700} & 
  422 & 58 & 137 & 1536 & 58 & 37 \\ 
  \textbf{RTX 5000} &  625 & 48 & 77 & 2857 & 47 & 16 \\ 
 \textbf{Loihi} &   50 & 0.42 & 8.4 & 97 & 0.24 & 2.47 \\ 
\hline \hline
  &   \multicolumn{3}{c|}{\textbf{Training}} & \multicolumn{3}{c|}{\textbf{Testing}} \\
\hline
\end{tabular}
\vspace{-3mm}
\end{table}

\subsubsection{Performance}

We experiment on four datasets; MNIST, Fashion-MNIST, CIFAR-10 and MSTAR. MNIST \cite{LeCun1998}, 
Fashion-MNIST \cite{Xiao2017} and CIFAR-10 \cite{krizhevsky2009learning} are standard benchmarks.
The MSTAR dataset \cite{darpa} is a collection of SAR (Synthetic Aperture Radar). The subset we use is the MSTAR/IU Mixed Targets with 10 classes of vehicles. The images are target chips taken from scenes of SAR images, where each chip is 128 by 128 which we center-crop the 64x64 and resize it to 32x32. For each dataset, we convert the pixels into rate coded spike trains of duration $T$.

Table \ref{tab:Performance} shows the inference results on MNIST, Fashion-MNIST, CIFAR-10 and MSTAR dataset for EMSTDP (both FA and DFA). 
We see that even with the constraints in Loihi we achieve respectable performance as compared to the full precision implementations. The drop in performance on Loihi can be attributed to the quantization error due to the limitation of 8 bit weights and computation in Loihi. We can also see that the DFA based system has slightly better accuracy. This is probably because the DFA skipped the hidden layers in the backward path and has less accumulated quantization errors.

\subsubsection{Power}\label{sec:power}

Table \ref{tab:Power} shows power and energy comparison with CPU (i7 8700) and GPU’s (RTX 5000) versions with batch size 1 and pretrained convolutional layers. Given that Loihi's maximum operating frequency is 10KHz and the SNN requires $T$ time steps to operate in each phase, each sample takes longer time than in CPUs and GPUs. The throughput is sufficient for online image processing. Reducing the duration of each phase will improve the throughput but also sacrifice the quality of learning. The active power consumed by Loihi running these networks are a several orders lower than CPUs and GPUs.
This can be attributed not only to the asynchronous and simplified architecture of Loihi but also the sparse nature of SNNs. As we are learning SNNs in-hardware, we compare the energy per sample for both training and testing time. 
During the inference mode, backward paths are not implemented, hence it has higher throughput and lower energy per image than the training model.

\begin{figure}[t]
  \centering
    \begin{subfigure}{0.45\textwidth}
    \centering
  \includegraphics[width=\linewidth,height=5cm,keepaspectratio]{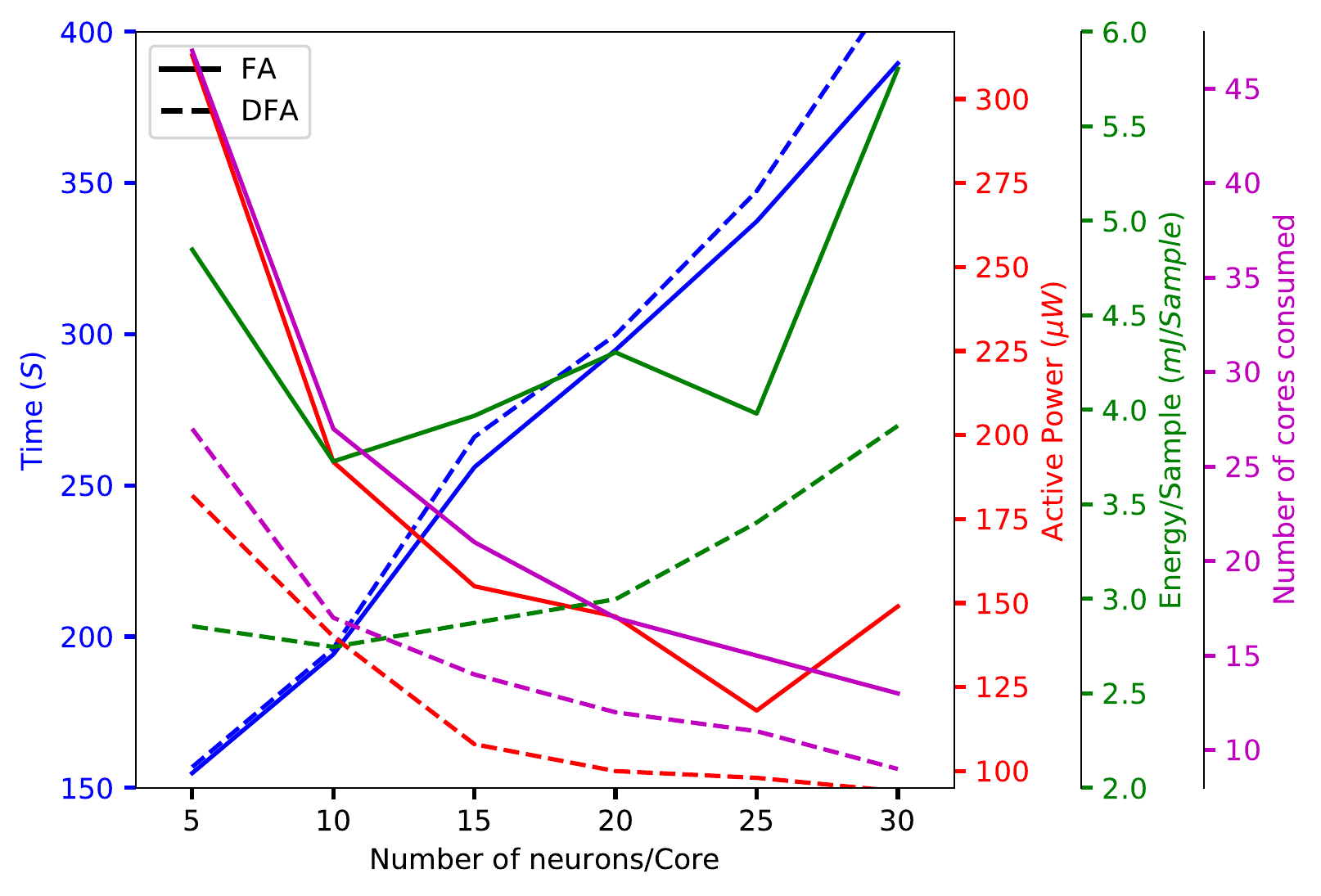}
  
  \end{subfigure}
  \vspace{-0.3cm}
  \caption{Trade-off between FPS and Active power consumption through Energy/Sample for EMSTDP on Loihi with FA and DFA while training with 10000 samples}
  \label{fig:power}
  \vspace{-5mm}
\end{figure}

In Fig.\ref{fig:power}, we show the effect of the iterative mapping of compartments (Section \ref{sec:map}) in the Loihi cores during training. When we increase the number of neurons mapped to each core, the number of occupied cored is reduced and the active power decreases as the cores that are not in use are power gated. At the same time the execution time increases as the core is shared by higher number of neuron compartments. As a result, the energy per sample first decrease and then increase. There is a best mapping size that minimizes the energy and it is necessary to analyze the trade-offs between the execution time and the power usage. For Table \ref{tab:Power}, we chose 10 logical neurons in the hidden layer to be packed in a single core based on Fig.\ref{fig:power}.

\subsubsection{FA versus DFA}

As we discussed in Section \ref{lab:ForErrPath}, utilizing DFA reduces the number of total compartments, cores and synapses. This effect can be seen from Fig. \ref{fig:power} where DFA based system consistently utilizes less cores, thus consumes less active power. Because the execution time is determined by the number of neurons per core, DFA and FA have similar throughput at the same neurons per core level. However, at the same performance level, DFA consumes lower power and dissipates less energy per image. 

\begin{figure}[t]
  \centering
    \begin{subfigure}{0.45\textwidth}
    \centering
  \includegraphics[width=\linewidth,height=4.5cm,keepaspectratio]{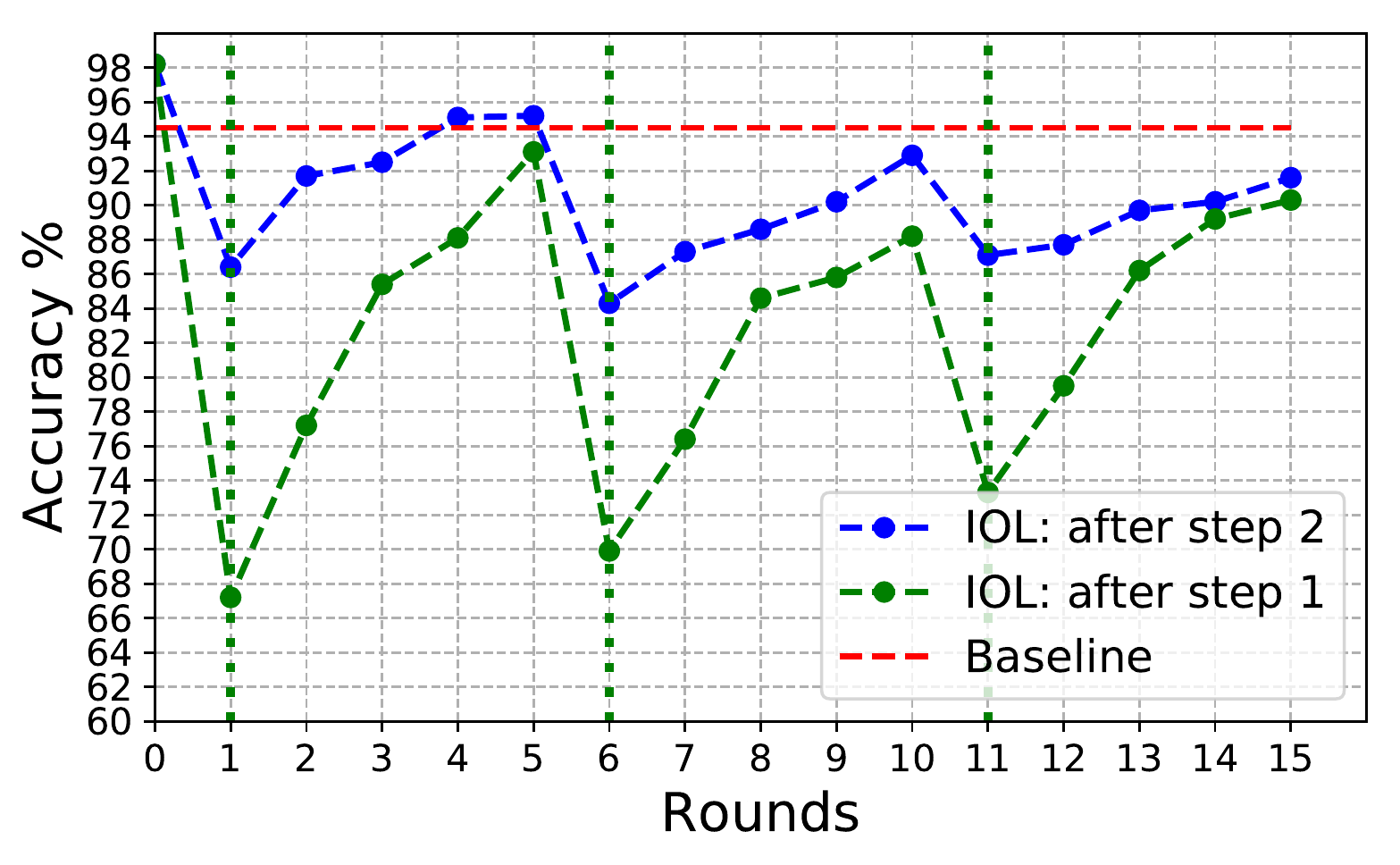}
  
  \end{subfigure}
  \vspace{-0.3cm}
  \caption{Incremental Online Learning with MNIST: 2 new classes are introduced at rounds marked by the green dotted line (Baseline: result from the same network trained with the entire dataset, IOL: Incremental Online Learning)}
  \label{fig:IOL}
  \vspace{-5mm}
\end{figure}

\subsection{Incremental Online Learning}

The in-hardware online learning provides the flexibility for the system to learn new classes after deployment or compensate possible model error due to poorly chosen training set or potential sensor hardware drifting. Such adaptability cannot be found in a system running models trained offline. In the second experiment, we demonstrate such flexibility using incremental online learning.

We set up our experiment to adopt an alternating two-step learning technique  similar to \cite{he2020incremental}; (1) learn new classes (2) retrain with new and old classes. A modified cross-distillation loss is used in step (1) to enable the model to minimize catastrophic forgetting which is a phenomenon where the performance on the old classes degrades dramatically as new classes are added. A cross-entropy loss is used in step (2) in which the retraining is done with the recently observed new classes and an equal size sample of old classes from a set with new observations of old classes as well. These new observations may have different distribution to that of old observations or could simply be noise or variations caused by the input device/sensor. This two step method helps to minimize concept drift \cite{he2020incremental} caused by these new observations in a standard online learning set up. 

For this experiment, we pretrain a model for 4 randomly selected classes in the MNIST dataset. Then we have three incremental training iterations, each time 2 new classes were added. We also divide ~6000 samples of each class into 5 chunks such that we can introduce the new classes over 5 rounds of learning new classes and retraining for each incremental training iteration. To approximate the effect of cross-distillation loss, we disable the classifier neurons of the old class and reduce the learning rate during step (1). In each round, sampling of old classes for step (2) is done from a set with both old and new observations of the old classes. Fig. \ref{fig:IOL} shows the accuracy of observed classes for all the incremental training steps at each round at the end of step (1) and (2) for the same network used for MNIST in Section. \ref{sec: OL}. As our method is only approximating cross-distillation, catastrophic forgetting can have a huge effect on our model's performance seen from the green dotted line in the beginning of each incremental training iteration. However, spreading the introduction of new classes over 5 rounds helps to alleviate that and recover. Thus, we see a big drop in accuracy in the first round of new classes and then recovery over the next rounds in each incremental learning step.

\section{Conclusion}

In this work, we adapted a spike-based backpropagation algorithm with biological plausible local update rules to fit the constraints of a neuromorphic hardware and implemented it on Intel’s Loihi chip. This resulting system enables low power in-hardware online supervised learning of multilayered SNNs which was tested on MNIST, Fashion-MNIST, CIFAR-10 and MSTAR datasets with promising performance and energy-efficiency. We also demonstrate  a  possibility  of  incremental  online  learning with  the  implementation.

\section*{Acknowledgement}
This work is partially supported by the National Science Foundation I/UCRC ASIC (Alternative Sustainable and Intelligent Computing) Center (CNS-1822165).

\bibliographystyle{IEEEtran}
\bibliography{main}

\end{document}